\documentclass[final]{cvpr}

\usepackage{times}
\usepackage{epsfig}
\usepackage{graphicx}
\usepackage{amsmath}
\usepackage{amssymb}
\usepackage{multirow}

\usepackage[symbol]{footmisc}

\usepackage{enumitem}

\usepackage[pagebackref=true,breaklinks=true,colorlinks,bookmarks=false]{hyperref}

\def\cvprPaperID{5450} 
\def\confYear{CVPR 2021}

\begin{document}

\title{CodedStereo: Learned Phase Masks for Large Depth-of-field Stereo}

\author{
Shiyu Tan\textsuperscript{1,*}
\qquad Yicheng Wu\textsuperscript{1,*}
\qquad Shoou-I Yu\textsuperscript{2}
\qquad Ashok Veeraraghavan\textsuperscript{1,$\dagger$}\\
\textsuperscript{1}Rice University\qquad
\textsuperscript{2}Facebook Reality Labs\\
{\tt\small \{shytan, yicheng.wu, vashok\}@rice.edu}
\qquad
{\tt\small Shoou-I.Yu@fb.com}
}
\maketitle
\thispagestyle{empty}
\pagestyle{empty}

\footnotetext[1]{These two authors contributed equally. \textsuperscript{$\dagger$}Corresponding author.}

\begin{abstract}
Conventional stereo suffers from a fundamental trade-off between imaging volume and signal-to-noise ratio (SNR) -- due to the conflicting impact of aperture size on both these variables.
Inspired by the extended depth of field cameras, we propose a novel end-to-end learning-based technique to overcome this limitation, by introducing a phase mask at the aperture plane of the cameras in a stereo imaging system.
The phase mask creates a depth-dependent yet numerically invertible point spread function, allowing us to recover sharp image texture and stereo correspondence over a significantly extended depth of field (EDOF) than conventional stereo.
The phase mask pattern, the EDOF image reconstruction, and the stereo disparity estimation are all trained together using an end-to-end learned deep neural network.
We perform theoretical analysis and characterization of the proposed approach and show a ~$6\times$ increase in volume that can be imaged in simulation. 
We also build an experimental prototype and validate the approach using real-world results acquired using this prototype system.

\end{abstract}
\vspace{-0.1in}

\section{Introduction}
\label{sec:intro}



Stereo-based 3D reconstruction, while extremely popular, suffers from a fundamental trade-off between volume of imaging and noise.
If you want to retain a large volume of imaging, then in order to get sharp texture features for correspondence, you need to ensure that the depth of field (DOF) of the cameras covers the entire volume. 
This necessitates a narrow aperture, rapidly reducing the total light level reaching the sensor (since SNR is quadratically related to aperture size). 
As a consequence, it is challenging to get large volume, high quality, and high resolution stereo-based 3D reconstruction in light-limited environments.

\begin{figure}
  \begin{center} 
  \includegraphics[width=\linewidth]{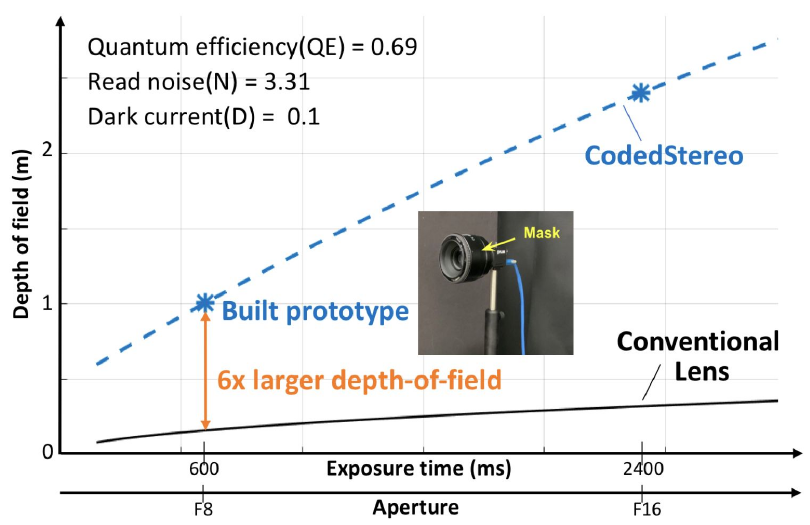}  
  \end{center}
  \caption{\textbf{Tradeoff between depth of field and aperture size on simulated data.} We propose a CodedStereo system that can provide an $6\times$ increase in DOF (blue dashed curve, with stars for particular observations). In the curves, the x-axis is linearly sampled in exposure time, and the corresponding f-numbers are converted to maintain the same SNR level of 50dB. Both our system and the conventional lens are focused at $1m$, with a $50mm$ focal length.}
  \label{fig:section3}
  \vspace{-0.1in}
\end{figure}

In light-limited environments, typically either the exposure duration or the aperture size of a camera is increased to increase light throughput. 
But for scenarios where there is either scene motion (eg., motion capture) or camera motion (eg., robotics, autonomous navigation), increasing exposure duration results in motion blur.
On the other hand, increasing the aperture size will result in a smaller depth of field, thereby reducing the volume that can be reconstructed.

Inspired by the extended depth of field (EDOF) imaging techniques \cite{dowski1995extended}, we present CodedStereo, a technique in which we add optimized phase masks to the aperture of each of the two stereo cameras.
These phase masks allow each camera to maintain a large aperture size, increasing the light throughput of the cameras.
Meanwhile, the phase masks are specially designed to produce a depth-dependent focal blur that allows back-end stereo algorithms to continue to retain high resolution and quality.
In addition, our learned phase mask not only enables more accurate depth estimation, but also encourages sharper extended depth of field RGB images that can be used for downstream applications such as point cloud colorization. The reconstruction algorithms and the phase masks are simultaneously optimized using an end-to-end learning framework.
The main technical contributions of this paper are:

\renewcommand{\labelenumi}{\Roman{enumi}.}
\begin{enumerate}[nolistsep]

  \item We propose CodedStereo, a technique to recover large-volume, high-quality, and high-resolution 3D reconstructions in light-limited environments. The key idea in CodedStereo is the introduction of a phase mask in the aperture of the stereo cameras that allows us to increase the aperture size of the cameras without sacrificing the depth of field.
  \item We develop an end-to-end learning framework to jointly optimize the phase masks and the algorithms both for RGB image and disparity reconstructions.
  \item We demonstrate the significant performance benefits of CodedStereo both in simulation and using a prototype system.
 \end{enumerate}

\section{Related work}
\label{sec:related}
\textbf{Stereo matching.}
Given two (or more) cameras looking at the scene from different perspectives, the goal of stereo is to find the corresponding scene points between the two camera views and use this to estimate depth based on triangulation.
Traditional methods\cite{scharstein2002taxonomy, hirschmuller2007stereo} typically formulate it as a multistage optimization problem, including matching cost computation, cost aggregation, disparity optimization, and post-processing.
Recently, learning-based stereo algorithms have become popular primarily due to their improved performance. 
Many networks, inspired by the traditional stereo matching pipelines, have been shown to achieve state-of-art results \cite{pang2017cascade, chang2018pyramid, mayer2016large, duggal2019deeppruner, zhang2019ga}.
Among these algorithms, \cite{mayer2016large, duggal2019deeppruner} are computationally efficient and can be used for real-time inference. 
However, it is well known that existing stereo algorithms degrade in performance when the images contain significant blur or noise \cite{liu2020visually, jeon2016stereo}.


\textbf{Low light stereo.}
Extending stereo algorithms and improving their performance in the presence of significant noise (as is the case in low-light imaging) is an area of active research.
The simplest solutions attempt to first denoise the stereo pairs before the correspondence search.
But unlike generic denoising algorithms \cite{chen2018learning,liba2019handheld}, these methods pay more attention to the consistency of the denoised image pairs to make sure the stereo matching algorithms can find the corresponding features. 
Another technique to improve low-light performance is to replace one or both of the stereo cameras with monochrome sensors, resulting in approximately a $3\times$ increase in light throughput \cite{jeon2016stereo}.

\textbf{Stereo and defocus blur.}
When the camera aperture is large such that the scene is no longer contained within the depth of field of the camera, focus blur is apparent in the captured images.
There have been attempts to exploit this focus blur as an additional depth cue to compensate for the degraded stereo performance \cite{chen2015blur,wang2016depth,chen2017learning,gil2019monster}.
Furthermore, Takeda \etal~\cite{takeda2013fusing} proposed the addition of amplitude masks in the aperture plane. The use of amplitude masks increases the variations in the depth-dependent blur, improving depth from defocus approaches. 
Our technique also proposes the addition of a mask within the camera's aperture plane. However, there are two key differences. 
First, since our main goal is low-light imaging, we use phase masks instead of amplitude masks to obtain high light throughput. 
Second, compared to the heuristic mask design in ~\cite{takeda2013fusing}, the proposed design is directly optimized based on the 3D reconstruction, which improves the performance.

\textbf{Extended depth of field imaging.}
For a conventional camera, it is well understood that the aperture size controls the relationship between the depth of field and SNR.
Larger apertures result in higher light collection leading to an increase in SNR, but at the cost of decreasing the depth of field.
There have been a host of techniques that have been developed to maintain a large aperture and a large depth-of-field.
One idea that has emerged from this line of inquiry is to reconstruct all-in-focus images from integrated images with a shaking sensor \cite{nagahara2008flexible}.
Another key idea is the use of a phase mask in the aperture plane to control the depth-dependent blur in a manner that makes the resultant blur invertible \cite{dowski1995extended, cossairt2010spectral, cossairt2010diffusion, elmalem2018learned, sitzmann2018end}.
Our design is intimately related to these efforts and the main difference is that when applying these techniques to stereo, one must pay attention to maintaining consistency across views, so that correspondence matching algorithms remain stable.

\textbf{End-to-end mask design.}
Over the last few years, several techniques have emerged where optical system design parameters and reconstruction algorithms are jointly optimized in an end-to-end manner. 
The primary rationale for this end-to-end learning framework is the significant improvements that are obtained as a result of this joint optimization.
Such methods have been shown to achieve superior performance in demosaicing \cite{chakrabarti2016learning}, monocular depth estimation \cite{wu2019phasecam3d,chang2019deep,haim2018depth},   microscopy \cite{nehme2020deepstorm3d, jin2020deep}, structured light \cite{baek2020polka,wu2020freecam3d}, EDOF \cite{sitzmann2018end}, and high dynamic range \cite{metzler2020deep,sun2020learning} imaging.
Our technique is of a similar vein, but tackling the problem of large volume, low-light stereo reconstruction.

\section{Imaging Volume vs SNR: The tradeoff}
\label{sec:tradeoff}


Traditional stereo exhibits a fundamental trade-off between light level, exposure time, and volume of reconstruction that limits the quality of 3D reconstructions.
In a traditional camera, the image signal to noise ratio (SNR) is proportional to incident light intensity, which in turn is proportional to the product of aperture area, illumination level, and exposure duration. Thus, 
\setlength{\abovedisplayskip}{4pt}
\setlength{\belowdisplayskip}{4pt}
\begin{equation}
    \textrm{SNR} \propto \frac{L_s TD^2}{\sigma_{tot}} = \frac{L_sT f^2}{\sigma_{tot}F_\#^2}.
\end{equation}
\noindent where $D=f/F_\#$ is the aperture diameter, $f$ is the focal length, $F_\#$ is the f-number, $T$ is the exposure duration, $L_s$ denotes the average light intensity, and $\sigma_{tot}$ refers to the total noise. This relationship indicates that the imaging SNR will become extremely low either under low-light conditions or when scene/camera dynamics require the use of short exposure durations.

Imaging SNR, in turn, impacts the quality of correspondence between the stereo pair, resulting in low-quality and/or low-resolution 3D reconstructions. The simplest solution to improve imaging SNR is to use a larger aperture. Unfortunately, this is typically not a feasible solution for stereo systems since this reduces the depth of field, which, as a result, significantly reduces the imaging volume for accurate 3D reconstruction. Specifically, the approximate depth of field (DOF) can be determined by focal length $f$, distance to subject $z_0$, acceptable circle of confusion size $c$, and aperture diameter \cite{jacobson2000manual}.
\setlength{\abovedisplayskip}{4pt}
\setlength{\belowdisplayskip}{4pt}
\begin{equation}
    DOF \propto \frac{2z_0^2F_\#c}{f^2}.
\end{equation}
\quad Figure \ref{fig:section3} demonstrates this tradeoff between depth of field and aperture size (or F\#) for a lens with a focal length of $50 mm$ and focused $1 m$ in front of the lens.
This shows that if you want a large imaging volume, then you need to use a narrow aperture. 
In particular, achieving a DOF of $1 m$ requires the use of a $F32$ aperture. Unfortunately, under low-light conditions and/or with short exposure duration, such a small aperture size would severely limit light throughput resulting in extremely noisy images -- which in turn will result in low-quality 3D reconstructions.
On the other hand, a large aperture in search of better light throughput induces significant focus blur within the imaging volume. This blur again affects the quality of stereo correspondences and 3D reconstruction.
Inspired by the extended depth of field imaging techniques, the key idea in our approach is to utilize a phase mask at the aperture plane to 1) keep the aperture large, and 2) create depth-dependent yet numerically invertible focal blur point spread functions that allow for high-quality 3D reconstruction over the entire imaging volume.
The increase in imaging volume/DOF achieved by CodedStereo is shown for comparison in Figure \ref{fig:section3}.


\section{Extended depth of field in stereo matching}
\label{sec: EDOF}


\begin{figure}
  \begin{center} 
  \includegraphics[width=\linewidth]{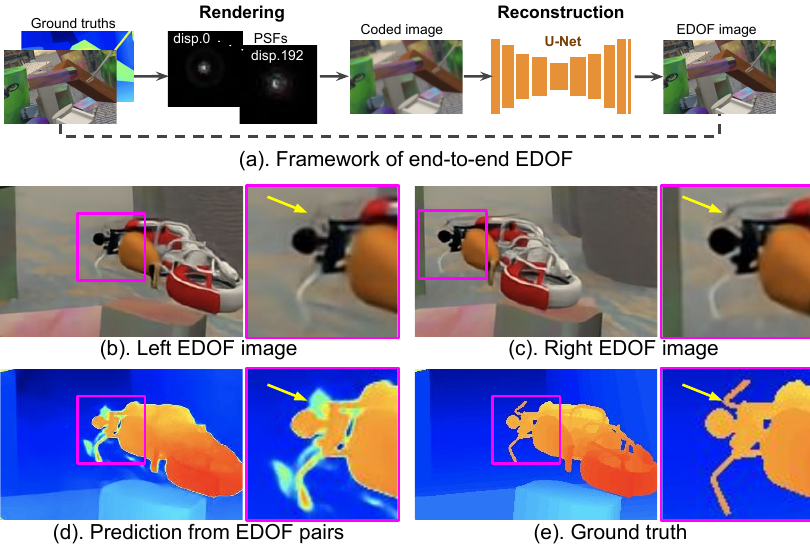}  
  \end{center}
  \vspace{-0.05in}
  \caption{\textbf{Na\"ive EDOF stereo results in 3D reconstruction artifacts due to feature inconsistencies.} (a) The framework used to learn the e2eEDOF phase mask. (b)-(c) Reconstructed EDOF images with inconsistent fine features across views. (d) Predicted disparity map from EDOF pairs. The algorithm failed to recover stereo correspondence for unmatched regions. (e) Ground truth.}
  \label{fig:EDOF}
  \vspace{-0.05in}
\end{figure}

\begin{figure*}
  \begin{center}
  \includegraphics[width=\linewidth]{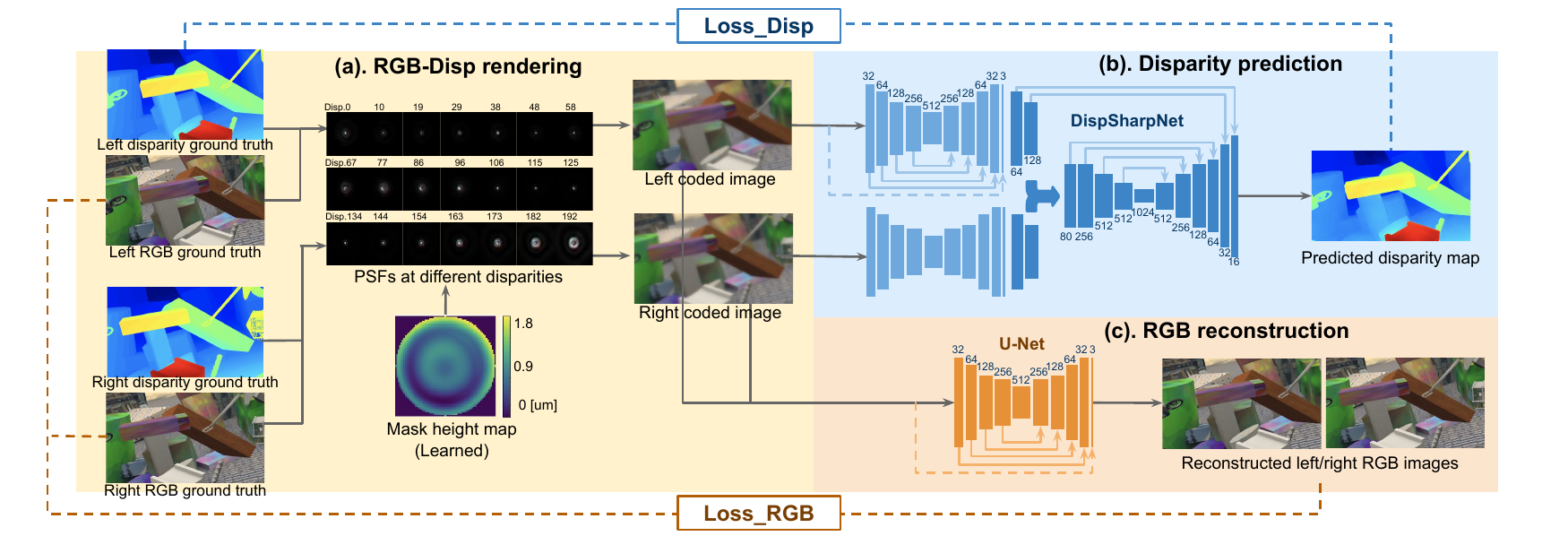}
  \end{center}
  \caption{\textbf{Framework overview}. We learn the phase mask together with a disparity prediction network and an RGB reconstruction network in an end-to-end manner. 
  In the RGB-Disp rendering layer, disparity-dependent PSFs are first simulated given the learnable phase mask. These PSFs are then convolved with ground truths to render left/right coded images, which are the inputs to the following reconstruction networks. We use a DispNet-based network and a U-Net-based network to estimate the sharp texture images and the disparity map, respectively. The loss of reconstructed texture and disparity are summed up together in backpropagation to update the mask height map and the network parameters at the same time.}
  \label{fig:pipeline}
\vspace{-0.1in}
\end{figure*}


One na\"ive technique to overcome the imaging volume vs SNR tradeoff in stereo systems would be to replace each of the cameras in a stereo system with an EDOF camera.
Surprisingly, this na\"ive application of EDOF does not seem to result in significant improvement to the imaging volume in stereo systems.
The primary reason for this discrepancy is that the deconvolution algorithms, irrespective of whether they are optimization-based \cite{dowski1995extended, cossairt2010spectral, cossairt2010diffusion, nagahara2008flexible} or learning-based \cite{sitzmann2018end, elmalem2018learned}, produce minor inconsistencies across views. 
While these inconsistencies are imperceptible and do not seem to affect the perceptual quality of individual images, they have a significant effect on the stereo correspondence search.
As a result, the stereo correspondence search produces significant errors, affecting the quality of the 3D reconstructions.

Figure \ref{fig:EDOF} shows the effect of these imperceptible inconsistencies on stereo reconstructions.
We follow the technique in \cite{sitzmann2018end} to learn an optimal phase mask for EDOF imaging and use that same phase mask for both the left and the right camera in a stereo system.
The e2eEDOF learning framework and the prediction results from EDOF pairs are shown in Figure \ref{fig:EDOF}. 
A close inspection of the results shows that the matching algorithm failed to recover correspondence due to the inconsistencies in the individual EDOF recovered images, as pointed to by the yellow arrow.


\section{CodedStereo framework}
\label{sec:framework}
Our technique consists of a single optimized phase mask inserted into the aperture of both the cameras in a stereo pair.
With these phase masks inserted, the aperture of these cameras can remain wide open, allowing significantly larger light collection thereby improving imaging SNR. 
The depth-dependent blur caused by the insertion of these phase masks is jointly optimized along with the disparity and image reconstruction algorithms to maximize the volume-SNR tradeoff in stereo. 
We call our technique 'CodedStereo'. 
Our system simultaneously obtains sharp image texture and stereo correspondence in a large depth of field, without sacrificing SNR or light throughput.

As shown in \ref{fig:pipeline}, the end-to-end training pipeline consists of three distinct parts: (a) \textit{Rendering:} A RGB-Disp simulator to render left/right coded images using texture and depth as inputs (while accounting for the depth-dependent defocus effect of a particular phase mask), (b) \textit{Disparity Prediction:} a DispNet-based deep network to estimate disparity from coded pairs, and (c) \textit{RGB Image Reconstruction:} a U-Net to reconstruct sharp images. 
The detailed description of each component is discussed in the following subsections.

\subsection{Rendering Using RGB-Disp Simulator}
In conventional stereo, it is assumed that the entire scene is within the depth of field.
When this is not true, as is the case here, the defocus blur apparent on the captured images depends upon the depth of the scene point, and thus depends upon the disparity between the corresponding points of two camera views.
In addition, when a phase mask is inserted into the aperture plane, the disparity-dependent point spread function (PSF) also depends upon the phase mask pattern.
The goal within the rendering layer is to accurately model the effect of phase mask pattern and disparity on the captured left and right images in a CodedStereo system.

The RGB-Disp rendering is based on Fourier optics theory~\cite{goodman2005introduction} and is fully differentiable to enable end-to-end training. 
We first simulate the point spread functions (PSFs) for each camera as the squared magnitude of the Fourier transform of the pupil function (which depends on the phase mask pattern)
\setlength{\abovedisplayskip}{3pt}
\setlength{\belowdisplayskip}{3pt}
\begin{equation}
    PSF_{\lambda}\propto|\mathcal{F}\{ A\exp(\phi^{M}+\phi^{DF}) \}|^{2},
\end{equation}  

\noindent where $\lambda$ is the wavelength. 
In the pupil function, $A$ denotes a circular amplitude function with respect to aperture size, $\phi^{M}$ denotes the phase shift induced by the phase mask (proportional to the mask height map), and $\phi^{DF}$ denotes the defocus phase. 
The defocus phase can be further derived as a function of disparity $d$,
\setlength{\abovedisplayskip}{4pt}
\setlength{\belowdisplayskip}{4pt}
\begin{equation}
\phi^{DF}(x_1,y_1)=\frac{k_\lambda}{2fb}(d-d_0)(x_1^{2} + y_1^{2})
\label{DF}
\end{equation}

\noindent where $k_\lambda=2\pi/\lambda$ is the wavenumber, $f$ is the focal length, and $b$ is the baseline between two views. 
$(x_1,y_1)$ denotes the spatial coordinate on mask plane, and $d_0$ is the corresponding disparity value at in-focus depth. 
We then render the coded images by convolving the ground truth RGB textures with the disparity and wavelength-dependent PSFs. 
\setlength{\abovedisplayskip}{5pt}
\setlength{\belowdisplayskip}{0pt}
\begin{equation}
    I^{c}_\lambda=\sum_d M_d\cdot(I^{i}_{\lambda}\ast PSF_{\lambda,d})+ noise
\end{equation}
\noindent where $\cdot$ is an element-wise product operator, $I^{i}$ is the input all-in-focus image, and $I^{c}$ is the rendered coded image. $M_d$ denotes a segmentation mask ($1$ when the pixel disparity is $d$, $0$ otherwise). To account for boundary occlusions, the segmented layers were further blended using the normalized matting weights\cite{lee2008real}. 
To render the effect of noise (which would be significant under low-light conditions), we apply an additive Gaussian noise, whose standard deviation is calculated based on the aperture size, light-level, and exposure duration.


\subsection{RGB and Disparity Reconstruction}
We use two separate networks to reconstruct the disparity map and sharp texture images. 
The texture reconstruction network is based on a modified residual U-Net~\cite{ronneberger2015u} in which the differences between coded image and ground truth image (i.e. residual image) are learned. 
The advantage of learning a residual image is to encourage high-frequency information recovery, like edges and detailed textures, and therefore such residual learning techniques are widely used in per-pixel estimation problems such as deblurring.

For disparity prediction, we adopt the  structure of DispNetC \cite{mayer2016large}. Note that DispNetC only outputs disparity maps at half the resolution of the input stereo pairs. 
We modify it by adding extra deconvolution layers to upsample the disparity map \cite{pang2017cascade}, so that the final output is the same size as the input left/right coded images. We further found that an extra encoder-decoder module before DispNetC can benefit the feature extraction of stereo pairs especially in areas of the image with a large amount of out-of-focus blur. 
Therefore, we process coded left/right images separately through a shared-weighted encoder-decoder layer followed by two convolution layers to extract features, and then horizontally correlate the features. We considered a maximum shifting of 64 pixels which corresponds to 192 pixels in the original coded images. 
We call our disparity prediction network DispSharpNet, as it enables disparity estimations with extra details and sharper boundaries. 
More details of network architectures are shown in the supplementary.




\begin{figure}
  \begin{center}
  \includegraphics[width=\linewidth]{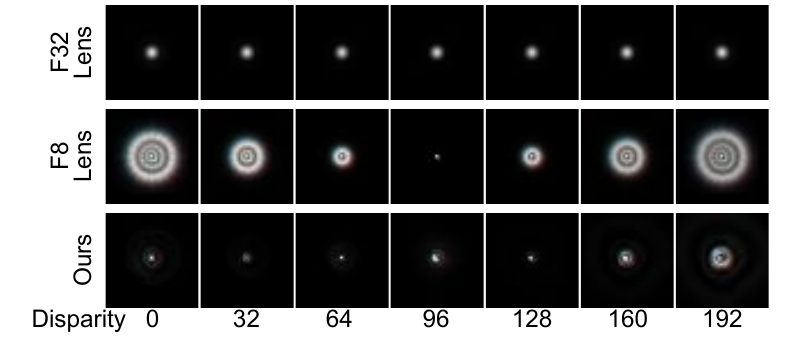}
  \end{center}
  \vspace{-0.05in}
  \caption{\textbf{Optimized PSFs in simulation (focus at 96px).} Each PSF slice is normalized for better visualization. PSFs for conventional $F32$ and $F8$ lenses are also shown for comparison.}
  \label{fig:sim_PSFs}
  \vspace{-0.1in}
\end{figure}

\subsection{Implementation details}
\label{sec:implmentation_details}
We optimized the phase mask over a depth range of [$0.7m-1.7m$] for a stereo system with a baseline of $22 mm$.
The lenses are focused at $1m$ with focal lengths of $50mm$ and $F8$ aperture sizes.
The sensors' pixel size was set to $4.8\mu m$, corresponding to a disparity range of [$134-326$] pixels.
To avoid large disparity values, we manually pre-shifted the right image by $134$ pixels to the right. This is equivalent to reduce the disparities by $134$, and thus the disparity range changes to [$0-192$].
During training, the mask was directly optimized over the reduced disparities (21 distinct values sampled in $[0-192]$).
The learnable mask height map was discretized with a pitch size of $88 \mu m$ at a resolution of $71\times 71$. Similar to the previous works \cite{shechtman2014optimal, wu2019phasecam3d}, we further parameterized the height profile and represented it using Zernike polynomials with $55$ coefficients.

\textbf{Loss function.} During training, the loss function is defined as a combination of disparity prediction error and RGB reconstruction error. 
We made use of the root mean squared error for both the estimated RGB image $\hat{I}$ and the predicted disparity $\hat{d}_i$ at different resolutions $i$.
\setlength{\abovedisplayskip}{4pt}
\setlength{\belowdisplayskip}{4pt}
\begin{equation}\footnotesize
\begin{split}
& Loss=Loss\_Disp + Loss\_RGB\\
& = \frac{1}{\sqrt{M}}
    \sum_i{\alpha_i\left \|d_i - \hat{d_i} \right \|_2}
    + \frac{1}{\sqrt{N}}\gamma(\left \|I^{l} - \hat{I}^{l} \right \|_2 + \left \|I^{r} - \hat{I}^{r} \right \|_2),
\end{split}
\end{equation} 
\noindent where $\alpha_i$, $\gamma$ are the corresponding weights, and $M$, $N$ are the number of pixels in the RGB image and disparity map, respectively. $l$ denotes the left, and $r$ denotes the right.  For stable feature matching, similar to DispNet \cite{mayer2016large}, we adopted a loss weight schedule to start training with only the lowest resolution loss, and progressively increase the weights of losses with higher resolutions.

\textbf{Dataset \& training.} Our model was end-to-end trained on a synthetic dataset consisting of dense ground truth disparity maps (enabling our RGB-Disp rendering) for $35{,}454$ training and $4370$ testing stereo pairs \cite{mayer2016large}. 
During training, the image patches were randomly cropped into a size of $384\times 768$, and preprocessed by subtracting out their means and dividing by their standard deviations. 
We optimized our phase mask and network parameters using Adam optimizer ($\beta_1=0.9$, $\beta_2=0.999$) with a batch size of $8$ for $50$ epochs, on GeForce RTX 2080 Ti GPUs.

\begin{figure*}
  \begin{center} 
  \includegraphics[width=\linewidth]{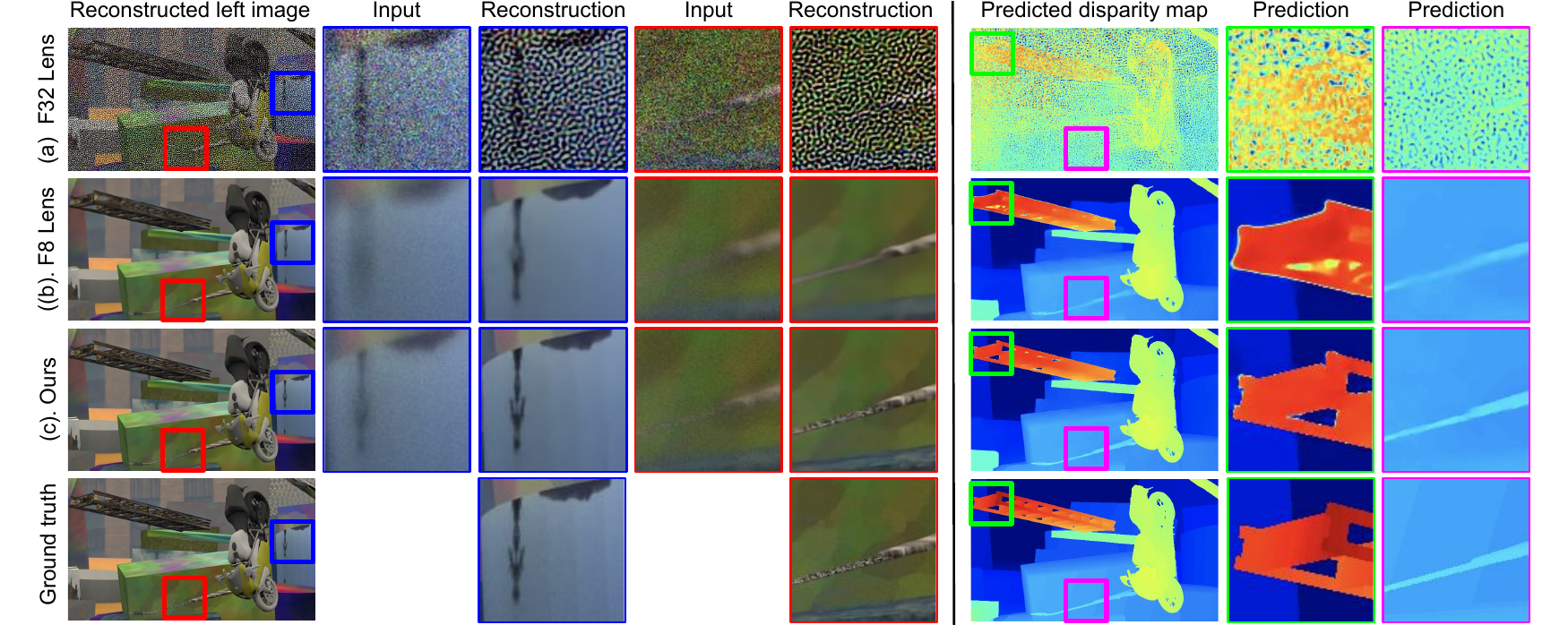}
  \end{center}
  \caption{\textbf{Comparison with conventional baselines (in simulation)}. (a) with small-aperture conventional $F32$ lenses. (b) with open-aperture conventional $F8$ lenses. (c) ours with optimized masks.
  For comparison, we applied the same reconstruction networks to $F32$ and $F8$ systems as ours, i.e. U-Net for RGB images estimation, and DispSharpNet for disparity prediction. Results show that our design outperforms conventional designs in a high-quality, high-resolution reconstruction with clear details and sharp edges.}
  \label{fig:sim_results}
  \vspace{-0.05in}
\end{figure*}

\section{Simulation results}
\label{sec:sim}
\begin{figure}
  \begin{center}
  \includegraphics[width=\linewidth]{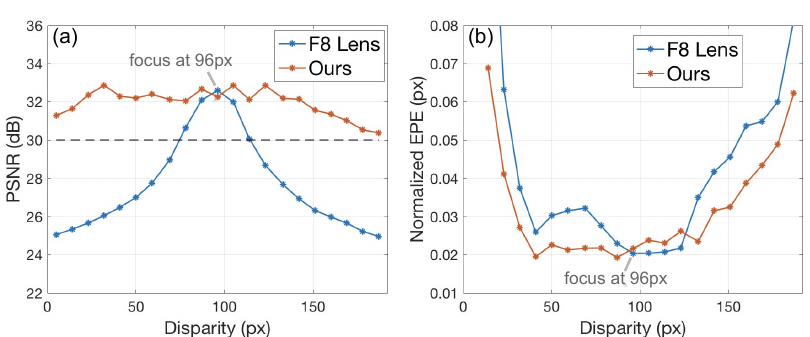}
  
  \smallskip
  \small
  \begin{tabular}{c|c|c|c|c}
  \hline
  \multicolumn{2}{c|}{}
  & F32 Lens& F8 Lens & Ours \\
  \hline
  \multirow{2}{*}{\rotatebox[origin=c]{90}{RGB}}
  & PSNR[dB] & 11.27 & 28.52 & \textbf{31.90} \\
  & SSIM & 0.048 & 0.807   & \textbf{0.880} \\
  \hline
  \multirow{2}{*}{\rotatebox[origin=c]{90}{Disp.}}
  & EPE[px] & 38.034 & 1.815 & \textbf{1.512}   \\
  & 3px[\%] & 95.45\%& 9.79\%  & \textbf{7.85\%}   \\
  \hline
  \end{tabular}
  \end{center}
  \caption{\textbf{Comparison with conventional lenses}.
  Top: The reconstructed PSNR and EPE (normalized to disparity ground truth) variations with disparity are plotted. Our method is significantly better than conventional baselines, especially at the out-of-focus range, resulting in a $6\times$ increase in depth of field (black dashed line for PSNR threshold at 30dB).
  Bottom: Average PSNR and SSIM are used for evaluations on texture reconstruction (the higher the better), and average EPE and 3-pixel error rate are used for evaluations on disparity prediction (the lower the better).}
  \vspace{-0.1in}
  \label{fig:sim_baseline}
\end{figure}

We conducted quantitative and qualitative evaluations of our method in simulation. The phase masked learned with $\gamma$=$0.5$ was selected for evaluations, both in simulation and in experiment, as it simultaneously produces the sharp RGB texture and accurate disparity map over a large depth of field. The optimized PSFs are shown in Figure \ref{fig:sim_PSFs}. Compared to a conventional $F8$ lens, our PSFs have a significantly shrunken radius of the Airy disk (comparable or even smaller than $F32$ lens) at out-of-focus depths, improving the reconstruction of both RGB images and disparity map with high resolution. Furthermore, our PSFs also come with some variations along the disparity axis, providing complementary blur cues to assist the disparity prediction of problematic areas.

\textbf{Comparison with conventional lenses}. 
To illustrate the improvement of our system over conventional designs, we compared our masks with a pair of $F32$ conventional lenses (small-aperture resulting in low SNR), and a pair of $F8$ lenses (open-aperture with a large amount of out-of-focus blur). For each system, the networks were trained with an additive $2\%$ Gaussian noise, assuming the cameras are all designed to work under normal-light conditions.

The average peak signal-to-noise ratio (PSNR) and the structural similarity (SSIM) are adopted for evaluations on the texture reconstruction, and the end-point error (EPE) and the 3-pixel threshold error rate (3px) are used for the disparity, as shown in Table \ref{fig:sim_baseline}. 
Our method outperforms conventional designs with higher RGB reconstruction accuracy and lower disparity prediction error.
A visual comparison is shown in Figure~\ref{fig:sim_results}. It is clear to see that the $F32$ system suffers from low SNR, resulting in noisy textures and disparity maps, while the $F8$ system fails to reconstruct fine features due to out-of-focus blur. Our design outperforms the $F32$ system and the $F8$ system in a high-quality, high-resolution reconstruction with clear details and sharp edges. We further compared the depth of field of the $F8$ system and ours, by analyzing the reconstruction PSNRs over disparities, as shown in Figure~\ref{fig:sim_baseline}. Our methods surpass the PSNR threshold (30dB) for all the disparities within the range [0-192], resulting in a $6\times$ increase in depth of field (invert disparity) compared to the $F8$ system.
The curves of the normalized disparity EPE (EPE divided by the ground truth) are also shown on the right, indicating our disparity prediction improvement in the out-of-focus range.

\begin{figure}
  \begin{center}
  \includegraphics[width=\linewidth]{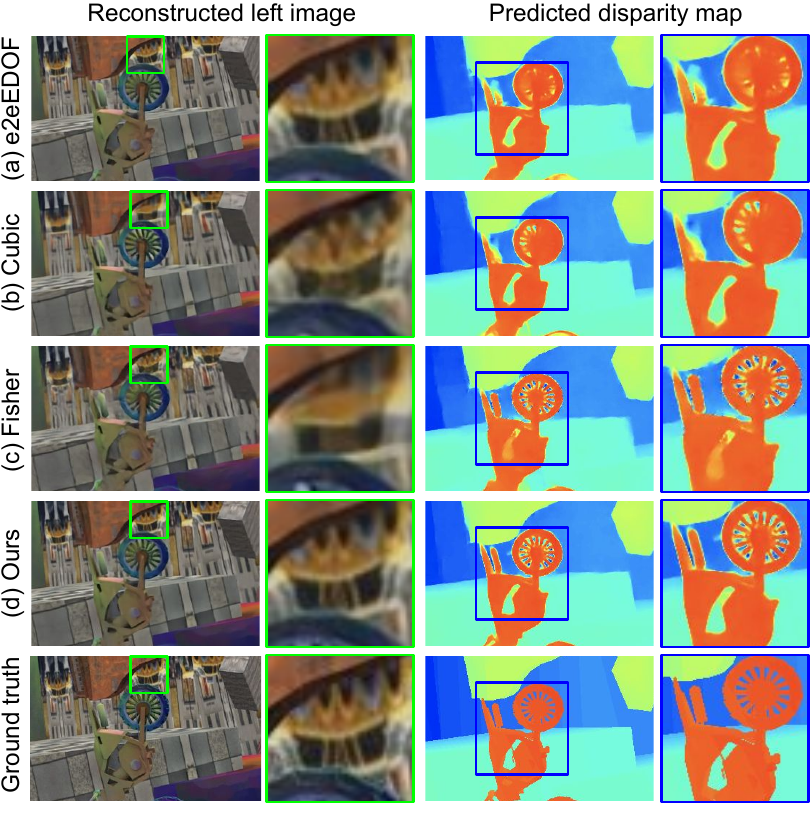}
  \end{center}
  \vspace{-0.06in}

  \small
  \begin{tabular}{c|c|c|c|c|c}
  \hline
  \multicolumn{2}{c|}{} 
      & e2eEDOF & Fisher & Cubic & Ours\\
  \hline
  \multirow{2}{*}{\rotatebox[origin=c]{90}{RGB}}
  & PSNR[dB] & \textbf{32.44} & 27.85 & 29.84 & 31.90 \\
  & SSIM & \textbf{0.880} & 0.820 & 0.839 & \textbf{0.880} \\
  \hline
  \multirow{2}{*}{\rotatebox[origin=c]{90}{Disp.}}
  & EPE[px] & 2.051 & 1.834 & 1.929 & \textbf{1.512}   \\
  & 3px[\%] & 11.71\% & 10.43\% & 10.28\% & \textbf{7.85\%}   \\
  \hline
  \end{tabular}
  \vspace{0.1in}
  \caption{\textbf{Comparison with other masks in simulation}. The e2eEDOF mask is end-to-end trained \cite{sitzmann2018end}, and its disparity is directly estimated from EDOF image pairs. The disparities of Fisher \cite{shechtman2014optimal} and Cubic \cite{dowski1995extended} masks are predicted from coded images. Our CodedStereo mask outperforms others on disparity estimation, and has comparable texture reconstruction accuracy to EDOF.}
  \label{fig:sim_com}
  \vspace{-0.1in}
\end{figure}

\textbf{Comparison with other masks.} We further compared our method with several other coded-aperture stereo systems. These coded masks were optimized based on the theoretical or heuristic properties of the PSFs. Specifically, the Fisher mask was designed to increase the PSFs variation over depth using Fisher information~\cite{shechtman2014optimal}, while the cubic mask was derived to force the PSFs to be similar over a large depth range \cite{dowski1995extended}. The comparison results are shown in Figure~\ref{fig:sim_com}. Reconstruction results of the e2eEDOF mask (Sec. \ref{sec: EDOF}) are also shown in the figure. Our optimized mask outperforms the e2eEDOF mask, the Fisher mask, and the cubic mask for both RGB and depth estimation.

\textbf{Ablation study.} As mentioned in Sec.~\ref{sec:implmentation_details}, the overall loss contains both the loss of RGB reconstruction and the loss of depth prediction, and $\gamma$ is the corresponding weight. In Table~\textcolor{red}{1}, we compared the performance under different $\gamma$ values. As expected, the network performs good depth estimation when $\gamma$ is small, and on the contrary, when $\gamma$ is large the network performs good RGB estimation. We finally chose $\gamma=0.5$ in our system.

\section{Real experiment}
\label{sec:exp}

\begin{figure}
  \begin{center}
  \includegraphics[width=\linewidth]{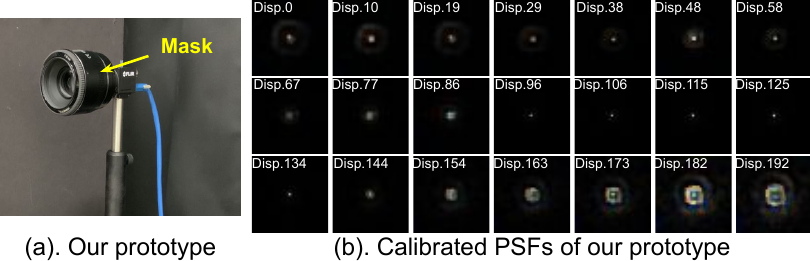}
  \end{center}
  \vspace{-0.05in}
  \caption{\textbf{Built prototype with calibrated PSFs.} We fabricated the mask and built a prototype to demonstrate our method.}
  \label{fig:exp_setup}
\vspace{-0.1in}
\end{figure}

To demonstrate our method, we built a hardware prototype with a fabricated mask inserted in a Yongnuo $50mm$ lens (with a $F8$ aperture). 
As shown in Figure \ref{fig:exp_setup}, a Blackfly (BFS-U3-200S6C-C) color camera with $2.4\mu m$ pixel size was used as the sensor. To match simulations, we sub-sampled the sensor pixels by $2\times2$ so that the equivalent pixel size is $4.8\mu m$ (with a resolution of $1824\times2736$). The left/right coded image pairs were captured by translating the camera $22 mm$ (baseline) using a Thorlabs linear stage. Similar to simulation settings, scenes were constructed within a volume of [$0.7m-1.7m$] from the prototype and the captured right images were pre-shifted by $134$ pixels to reduce the disparity value. The reduced disparity range then drops to [$0-192$], aligning with the settings for which the network was trained.
 
\textbf{Mask fabrication \& system calibration.}
We fabricated our mask using two-photon lithography (Photonic Professional GT Nanoscribe 3D printer). 
During printing, the height-map of the mask was discretized (in height) into $10$ steps with a stepsize of $200nm$. To account for any imperfection and misalignment in real experiments, we calibrated the PSFs with a deconvolution-based algorithm inspired by \cite{yuan2007image, wu2019phasecam3d}. The calibrated PSFs are shown in Fig.~\ref{fig:exp_setup}, which are used to finetune the reconstruction networks for best performance.
More fabrication and calibration details can be found in the supplemental material.  
 
\begin{figure*}
  \begin{center} 
  \includegraphics[width=\linewidth]{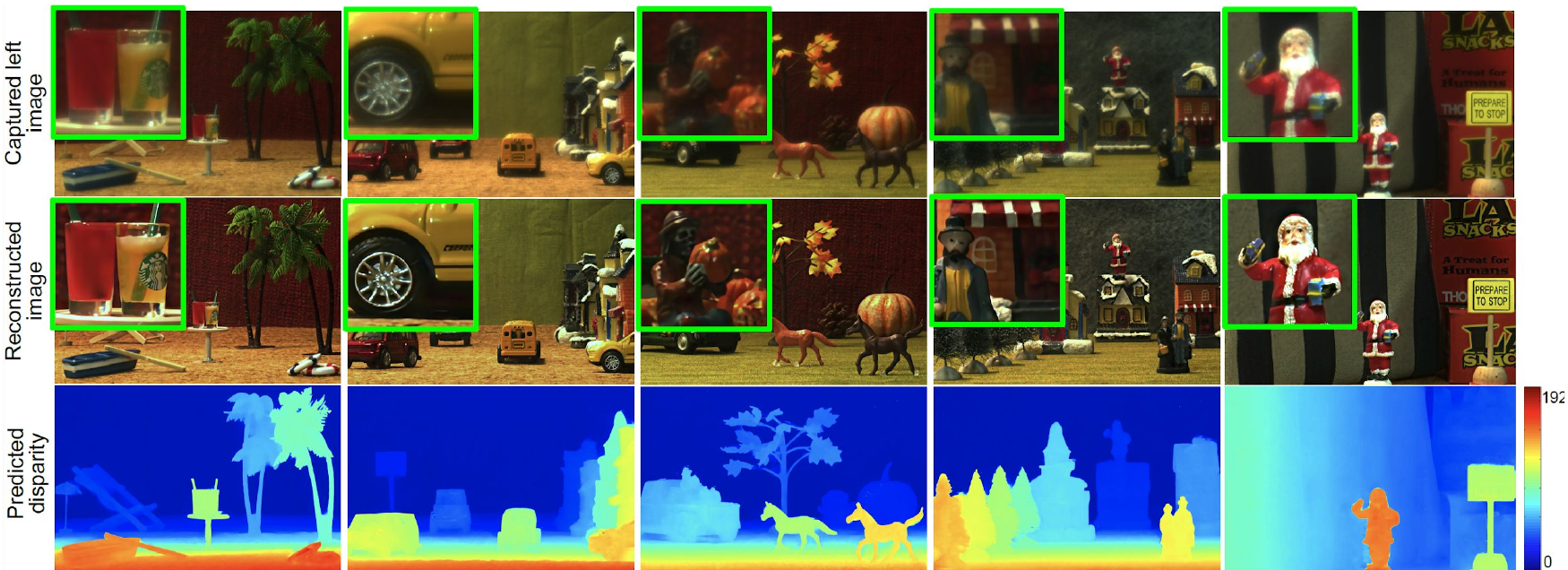}  
  \end{center}
  \vspace{-0.05in}
  \caption{\textbf{Experiment results of various real-word scenes using our CodedStereo prototype.} Reconstruction results are shown for real scenes with both uniform background and non-uniform background (the last column, variation in texture/depth).}
  \label{fig:exp_results}
\end{figure*}
\begin{figure*}
  \begin{center} 
  \includegraphics[width=\linewidth]{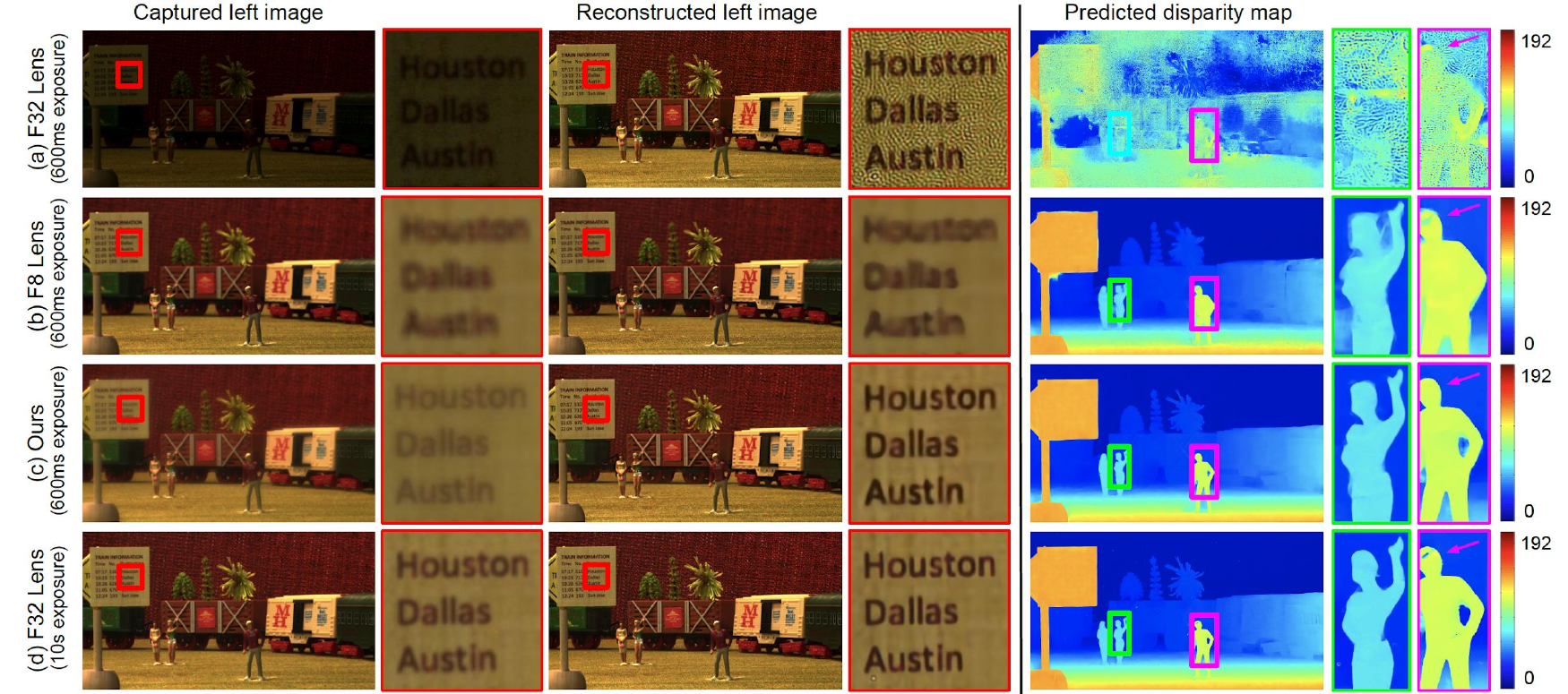} \end{center}
  \caption{\textbf{Comparison with conventional lenses in real-world experiments}. We compare the real-world performance of our prototype to the traditional $F32$ and $F8$ lenses here. The coded images of (a)-(c) are captured in the same 600ms exposure (scaled up by 8 times for $F32$ for visualization). (d) is long-exposure (10s) captured with a $F32$ lens, and the reconstructions are considered as the ground truths. As predicted by simulation, our system is superior to conventional designs in RGB and disparity reconstruction, and outperforms all these baselines (even long-exposure $F32$) on the disparity prediction in saturated regions, as pointed by the pink arrow.}
  \label{fig:exp_baselines}
  \vspace{-0.1in}
\end{figure*}

\textbf{Experiment results.}
Our real-world experiment results are shown in Fig.~\ref{fig:exp_results}. From the captured coded image pair, our method can reconstruct both RGB image and disparity with high accuracy in a large depth-of-range.
Similar to the simulation section, we further compared our prototype with conventional $F8$ and $F32$ lenses in real experiments. The same exposure time (600ms) was applied for all three settings. 
As a reference, we included the reconstruction results of a $F32$ system with 10s exposure time to show the best result we can get without the SNR constraint. The results are shown in Fig.~\ref{fig:exp_baselines}. 
As expected, the $F32$ system produces noisy reconstructions given low SNR, while the $F8$ system fails to recover fine features in texture and disparity due to the large out-of-focus blur. 
Our CodedStereo system generates high-quality results similar to the long-exposure $F32$ system with significantly shorter exposure time.

\begin{table}\small
\begin{center}
\begin{tabular}{c|c|c|c|c|c}
\hline
\multicolumn{2}{c|}{} 
      & $\gamma$=0 & $\gamma$=0.25 & $\gamma$=0.5 & $\gamma$=$\infty$\\
\hline
\multirow{2}{*}{\rotatebox[origin=c]{90}{RGB}}
 & PSNR[dB] & 28.82 & 30.34 & 31.90 & \textbf{32.44} \\
 & SSIM & 0.842 & 0.874 & \textbf{0.880} & \textbf{0.880} \\
\hline
\multirow{2}{*}{\rotatebox[origin=c]{90}{Disp.}}
 & EPE[px] & \textbf{1.462} & 1.477 & 1.512 & 1.718 \\
 & 3px[\%] &7.73\%& \textbf{7.25\%}& 7.85\%& 9.13\%\\
\hline
\end{tabular}
\label{tab:ablation_k1}
\end{center}
\caption{\textbf{Ablation study on various $\gamma$ values in loss function.} PSNR, SSIM of RGB reconstruction and EpE, 3-pixel error rate of disparity prediction as a function of $\gamma$.}
\end{table}

\section{Conclusion \& discussion}
\label{sec:con}

In this paper, we proposed a CodedStereo system that can recover large-volume, high-resolution 3D information under light-limited environments. The key idea of our system is to introduce a single phase mask at the aperture plane of stereo cameras. The mask was end-to-end learned together with an RGB reconstruction network and a disparity estimation network. The optimized phase mask creates a disparity-dependent point spread function, allowing us to recover sharp image and stereo correspondence over a significantly expanded depth of field than conventional stereo. We showed in simulation and experiments (with a prototype) that our method outperforms conventional lens and heuristic masks on both reconstructed texture and disparity.

Despite the advantages of our method, some limitations remain. First, the introduction of the phase mask makes the hardware system more complicated in design, and the re-training of phase masks and networks are required for different system settings (such as the lens focal length, the aperture size, the focus depth or the sensor pixel size that ends up with different defocus blur/disparities). 
Second, since our method is based on depth from disparity/defocus methods, it inherits their limitations on texture-less areas. Moreover, there is a trade-off between the accuracy of disparity and texture reconstruction (controlled by the weight $\gamma$). Further optimizing the system design might can mitigate this trade-off, including designs with two different phase masks/lenses across two views. Looking into the future, we hope to extend our framework to multi-view large depth-of-field stereo, enabling more reliable 3D information capturing under low-light conditions.

\noindent \textbf{Acknowledgement.} We would like to thank Hernan Badino and Matt Stewart at Facebook Reality Labs for their helpful discussions. 
AV, ST and YW were partially supported through a research grant from Facebook Reality Labs and the NSF CAREER Award \#1652633.

{\small
\bibliographystyle{ieee_fullname}
\bibliography{egbib}
}

\end{document}